\pdfoutput=1

\documentclass[11pt]{article}

\usepackage{emnlp2021}

\usepackage{times}
\usepackage{latexsym}
\usepackage{graphicx}
\usepackage{amsmath}
\usepackage{amssymb}

\usepackage[T1]{fontenc}

\usepackage[utf8]{inputenc}

\usepackage{microtype}

\usepackage{booktabs}
\usepackage{pgfplots}
\usepgfplotslibrary{groupplots}
\pgfplotsset{compat=1.16}

\usepackage{tikz}
\usepackage{tikz-qtree}
\usepackage{tikz-dependency}
\usepackage{booktabs}
\usepackage{algorithm}
\usepackage{algorithmic}

\title{Graph-Based Decoding for Task Oriented Semantic Parsing}

\author{Jeremy R. Cole$^\dagger$ \quad Nanjiang Jiang$^{\ddagger}$\thanks{\enskip Work done while on internship at Google.}  \quad Panupong Pasupat$^\dagger$  \quad Luheng He$^\dagger$  \quad Peter Shaw$^\dagger$  \\\\
 $^\dagger$Google Research \\
 {\tt \{jrcole,ppasupat,luheng,petershaw\}@google.com} \\
 $^\ddagger$The Ohio State University \\
 {\tt jiang.1879@osu.edu}\\}

\begin{document}
\maketitle

\begin{abstract}
The dominant paradigm for semantic parsing in recent years is to formulate parsing as a sequence-to-sequence task, generating predictions with auto-regressive sequence decoders. In this work, we explore an alternative paradigm. We formulate semantic parsing as a dependency parsing task, applying graph-based decoding techniques developed for syntactic parsing. We compare various decoding techniques given the same pre-trained Transformer encoder on the TOP dataset, including settings where training data is limited or contains only partially-annotated examples. We find that our graph-based approach is competitive with sequence decoders on the standard setting, and offers significant improvements in data efficiency and settings where partially-annotated data is available.
\end{abstract}

\section{Introduction}
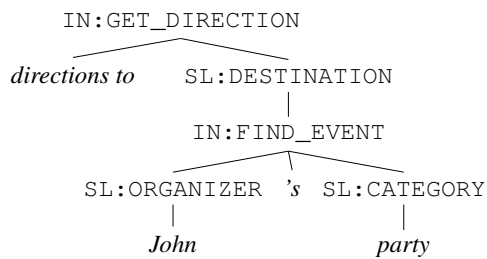
\begin{figure}[t!]
\centering
\scalebox{0.85}{
\begin{tikzpicture}
\tikzset{level distance=25pt, level 1/.style={sibling distance=-30pt}}
\Tree[.\texttt{IN:GET\_DIRECTION}
    {\emph{directions} \emph{to}}
    [.\texttt{SL:DESTINATION}
        [.\texttt{IN:FIND\_EVENT}
            [.\texttt{SL:ORGANIZER} {\emph{John}} ]
            {\emph{'s}}
            [.\texttt{SL:CATEGORY} {\emph{party}} ]
        ]
    ]
]
\end{tikzpicture}
}
	\caption{An example TOP~\cite{gupta-etal-2018-semantic-parsing} tree.}  
	\label{fig:top_example}
	\vspace{-1em}
\end{figure}

Semantic parsing, the task of mapping natural language queries to structured meaning representations, remains an important challenge for applications such as dialog systems. To support compositional utterances in a task oriented dialog setting, \citet{gupta-etal-2018-semantic-parsing} introduced the Task Oriented Parse (TOP) representation and released a dataset consisting of pairs of natural language queries and associated TOP trees. As illustrated in Figure~\ref{fig:top_example}, TOP trees are hierarchically structured representations consisting of intents, slots, and query tokens. 

We propose a novel formulation of semantic parsing for TOP as a graph-based parsing task, presenting a graph-based parsing model (hereafter, GBP). Our approach is motivated by the success of such approaches in dependency parsing~\cite{mcdonald2005online, kiperwasser2016simple, dozat-manning-2017-biaffine,kulmizev-etal-2019-deep} and AMR parsing~\cite{zhang-etal-2019-amr}.
Recently, sequence-to-sequence (seq2seq) models have become a dominant approach to semantic parsing (e.g., \citealt{dong-lapata-2016-language}, \citealt{jia-liang-2016-data}, \citealt{wang-etal-2019-rat}), including on TOP (e.g., \citealt{rongali-etal-2020-generate}; \citealt{aghajanyan-etal-2020-conversational}; \citealt{shao-etal-2020-graph}). Unlike such approaches that predict outputs auto-regressively, GBP decomposes parse tree scores over parent-child edge scores, predicting all edge scores in parallel.

First, we compare GBP with seq2seq and other decoding techniques, within the context of a fixed encoder and pretraining scheme: in this case, BERT-Base \citep{devlin-etal-2019-bert}. This allows us to isolate the role of the decoding method. We compare these models across the standard setting, as well as additional settings where training data is limited, or when fully annotated examples are limited but partially annotated examples are available. We find that GBP outperforms other methods, especially when learning from partial supervision. Second, we compare GBP with seq2seq models that additionally leverage pretrained decoders. We find that GBP remains competitive, and continues to outperform in the partial supervision setting.

\begin{figure*}[htb!]
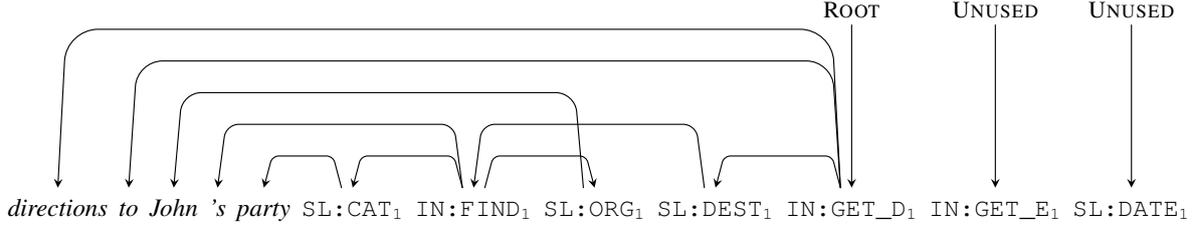

\centering

\scalebox{1.05}{
\begin{dependency}[theme = simple, segmented edge]
\tikzstyle{symbol}=[font=\small\ttfamily]
\tikzstyle{word}=[font=\small\itshape]  
\begin{deptext}[column sep=-1pt, nodes={symbol}, row sep=-3.0cm ]
 |[word]| directions \& 
 |[word]| to \& 
 |[word]|  John \& 
 |[word]|  's \& 
 |[word]|  party \& 
 SL:CAT\textsubscript{1} \& 
 IN:FIND\textsubscript{1} \& 
 SL:ORG\textsubscript{1}  \& 
 SL:DEST\textsubscript{1}  \& 
 IN:GET\_D\textsubscript{1} \& 
 IN:GET\_E\textsubscript{1} \& 
 SL:DATE\textsubscript{1} \& 
\\
\end{deptext}
\deproot[edge height=2.3cm]{10}{\large \textsc{Root}}
\depedge[edge height=2.0cm]{10}{1}{}
\depedge[edge height=1.6cm]{10}{2}{}
\depedge[edge height=1.2cm]{8}{3}{}
\depedge[edge height=0.8cm]{7}{4}{}
\depedge[edge height=0.4cm]{6}{5}{}
\depedge[edge height=0.4cm]{7}{6}{}
\depedge[edge height=0.4cm]{7}{8}{}
\depedge[edge height=0.8cm]{9}{7}{}
\depedge[edge height=0.4cm]{10}{9}{}
\deproot[edge height=2.3cm]{11}{\large \textsc{Unused}}
\deproot[edge height=2.3cm]{12}{\large \textsc{Unused}}
\end{dependency}
}

\caption{The graph-based model predicts parent assignments across a set of nodes consisting of query tokens, output symbols for intents and slots, and special \textsc{Unused} and \textsc{Root} symbols. This is the corresponding parse tree for the TOP tree shown in Figure~\ref{fig:top_example}. Not all output symbols are drawn; omitted symbols are attached to \textsc{Unused}. Intent and slot names are abbreviated.}
	\label{fig:model_parse}
\end{figure*}

\section{Task Formulation}

We present a novel formulation of the TOP semantic parsing task as a graph-based parsing task.
Our goal is to predict a TOP tree $\mathbf{y}$ given a natural language query $\mathbf{x}$ as input. The nodes in $\mathbf{y}$ consist of intent and slot symbols from a vocabulary of output symbols $\mathcal{V}$ and the tokens in $\mathbf{x}$. However, $\mathbf{y}$ cannot be predicted directly by a conventional graph-based approach~\cite{mcdonald2005online} for two reasons. First, given $\mathbf{x}$, we do not know the subset of intent and slot\footnote{Note that one could imagine instead treating slots as edge labels instead of nodes, but as the set is large (36 slots for 25 intents), little advantage would be expected.} symbols that occur in $\mathbf{y}$. Second, intent and slot symbols can occur more than once in $\mathbf{y}$.\footnote{See Figure~\ref{fig:repeated_example} in Appendix for an example.}

To address this, let us consider a parse tree $\mathbf{z}$ in a space of valid trees defined as $\mathcal{Z}(\mathbf{x})$. The parse tree $\mathbf{z}$ can be deterministically mapped to and from $\mathbf{y}$. The parse tree $\mathbf{z}$ consists of: (1) the tokens in $\mathbf{x}$, (2) every symbol in $\mathcal{V}$ replicated up to a maximum number of occurrences\footnote{The number of repetitions per output symbol is determined from the training data. If a symbol has a maximum of $k$ occurrences in a TOP tree in the training data, it will have $k+2$ replications. See Appendix~\ref{sec:repeated} for more information.} and assigned a corresponding index,
and (3) a special \textsc{Unused} node in addition to the standard \textsc{Root} node.
Let $\mathcal{N}(\mathbf{x})$ be this set of nodes which all trees in $\mathcal{Z}(\mathbf{x})$ consist of. When mapping from $\mathbf{y}$ to $\mathbf{z}$, output symbols occurring multiple times are indexed following a pre-order traversal, and any output symbol that does not occur in $\mathbf{y}$ is assigned to the \textsc{Unused} node in $\mathbf{z}$. For example, Figure~\ref{fig:top_example} illustrates an example TOP tree, $\mathbf{y}$, and Figure~\ref{fig:model_parse} illustrates a corresponding parse tree, $\mathbf{z}$.

\section{Scoring Model}
\label{sec:model}

Given that the mapping between $\mathbf{y}$ and $\mathbf{z}$ is deterministic, our goal is to model $p(\mathbf{z} \mid \mathbf{x})$. We follow a conventional edge-factored graph-based approach~\cite{mcdonald2005online}, decomposing parse tree scores over directed edges between parent and child node pairs $(p,c)$ in $\mathbf{z}$:
$$ p(\mathbf{z} \mid \mathbf{x}) = \prod_{(p,c) \in \mathbf{z}} \frac{\exp(\phi(p,c,\mathbf{x}))}{\sum_{p' \in \mathcal{N}(\mathbf{x})} \exp(\phi(p',c,\mathbf{x})) }, $$
where edge scores, $\phi(p,c,\mathbf{x})$, are computed similarly to the model of~\citet{dozat-manning-2017-biaffine}:
$$ \phi(p,c,\mathbf{x}) = (e^{p}_{\mathbf{x}})^{\intercal}Ue^{c}_{\mathbf{x}} + (e^{p}_{\mathbf{x}})^{\intercal}u, $$
where $e^{p}_{\mathbf{x}}$ and $e^{c}_{\mathbf{x}}$ are contextualized vector representations of the nodes $p$ and $c$, respectively, and $U$ and $u$ are a parameter matrix and vector, respectively.\footnote{For computational efficiency and to prevent invalid trees, we consider the score for assigning a token node as a parent for any other node to be a fixed value of $-\infty$.}

Node representations are computed differently for each node type in $\mathcal{N}(\mathbf{x})$. Encodings for token nodes are based on the output of a BERT~\cite{devlin-etal-2019-bert} encoder; replicated output symbols are embedded based on their symbol and index; \textsc{Root} and \textsc{Unused} nodes likewise have a unique embedding. All nodes are then jointly encoded with a Transformer~\cite{vaswani-etal-2017-attention} encoder,
which produces the contextualized node representations ${e_{x}}^{p}$ and ${e_{x}}^{c}$ which are used in the above equations to produce the factored edge scores. 

The scoring model is trained using a standard maximum likelihood objective.

\section{Parsing Algorithm}
\subsubsection*{Chu-Liu-Edmonds Algorithm}
The Chu-Liu-Edmonds (CLE) algorithm is an optimal algorithm to find a maximum spanning arborescence over a directed graph \citep{chu-1965-shortest,edmonds1965paths}. It has commonly been used for parsing dependency trees from edge-factored scoring models (e.g., \citealt{mcdonald2005online}; \citealt{dozat-manning-2017-biaffine}). Note that in an arborescence (hereafter tree), each node can have at most one `parent', or incoming edge. Thus, the algorithm first chooses the highest scoring parent for each node as the initial \emph{best parent}. It is possible these initial best parents already form a tree; however, it may instead produce a graph with cycles. In that case, CLE recursively breaks the cycles until the optimal tree is found. Note that CLE takes the index of the root of the tree as an input, and begins by deleting all incoming edges to enforce this constraint. Conventionally, in dependency parsing, the root of the tree is the special \textsc{Root} node.

This algorithm is optimal for dependency parsing; however, our formulation differs due to additional constraints based on how TOP trees are mapped to and from dependency trees. First, by convention, the parent of the \textsc{Unused} subtree must be \textsc{Root}. Second, the \textsc{Unused} subtree must be of depth 2: it cannot have any grandchildren. Finally, as valid TOP trees have only one root, the \textsc{Root} node must have only one `child', or outgoing edge. 

\subsubsection*{Unused Node Preprocessing}
As stated, our \textsc{Unused} subtree must only have depth 2 to follow our task formulation. Otherwise, the final tree score will be computed incorrectly when translating to a TOP tree, as the entire \textsc{Unused} subtree is effectively discarded. Thus, we first preprocess the \textsc{Unused} subtree to ensure depth 2. In practice, simply using the initial best parents will result in \textsc{Unused} subtrees with depth 3 or greater about 1\% of the time.

We resolve such cases by making a decision for each node \emph{a} whose initial best parent is \textsc{Unused} and has children itself. One option is to delete the edge to \emph{a} from \textsc{Unused}, making the next highest scoring edge the new best parent of \emph{a}. The cost of this action is equal to the difference in scores between the corresponding edges. Alternatively, we can take a similar action on each child of \emph{a}: delete the edge from \emph{a}, making the next highest scoring edge the new best parent. The cost of this action is equal to the difference in the corresponding edges summed over every child of \emph{a}. We iterate over the children of \textsc{Unused} that have children, selecting the action with the lower cost, until the constraint is met. Then, we no longer allow further modifications to the \textsc{Unused} subtree, effectively deleting it for the remaining stages of the algorithm.

Note that this algorithm is not necessarily optimal: the order in which we consider the children of \textsc{Unused} can affect the final result. However, we find this approximation to work well in practice.

\subsubsection*{Multiple Root Resolution}
Our second modification to the CLE algorithm concerns the \textsc{Root} node. Valid TOP trees are single-rooted: in our formalism, this means the \textsc{Root} node can only have a single child. To enforce this constraint, we want to choose the single child of \textsc{Root} that results in the highest scoring tree. We then provide this child's index to the CLE subroutine and delete all edges from \textsc{Root}, effectively discarding it. To find the best root, we start with the set of nodes whose initial best parent is the \textsc{Root} node. If this set is a singleton, we simply choose that node as the tree's root, providing its index to the CLE subroutine. In about 0.5\% of trees, there is more than one node. In that case, we run the CLE algorithm with each node as the given root index, taking the highest-scoring tree. This is still not guaranteed to be optimal: the optimal choice of the root node could have a different initial best parent than \textsc{Root}. However, this was not observed in our experiments and trying every node drastically increases the computation.

\section{Experiments}

\begin{table*}[htpb]
\begin{center}
\scalebox{0.9}{
\begin{tabular}{l ccc ccccc}
\toprule
& \multicolumn{3}{c}{Standard Supervision} & \multicolumn{5}{c}{Partial Supervision} \\
 \cmidrule(lr){2-4} \cmidrule(lr){5-9}
Decoder (BERT-Base encoder) & 100 & 10 & 1 & 10/90/0 & 10/0/90 & 1/99/0 & 1/0/99 & 2/49/49 \\ 
\midrule
\textsc{PtrGen} \citep{rongali-etal-2020-generate} & 83.13 & --- & --- & --- & --- & --- & --- & --- \\
\textsc{PtrGen} (our implementation)& 85.00 & 76.84 & 51.85 & 13.24 & 26.68 & \phantom{0}4.39 & \phantom{0}0.00 & \phantom{0}4.43 \\
\textsc{FSP} \citep{pasupat-etal-2019-span} & 85.12 & \bf 79.44 & \bf 57.95 & 68.94 & --- & 42.94 & --- & --- \\
\textsc{GBP} (proposed) & \bf 86.14 & 79.43 & 56.89 & \bf 84.55 & \bf 84.14 & \bf 81.52 & \bf 73.94 & \bf 85.01 \\
\bottomrule
\end{tabular}
}
\caption{Results for various decoders with BERT-Base as encoder. For \emph{standard supervision}, column headers denote the percentage of training data used. For \emph{partial supervision}, column headers S/T/N denote the percentage of training examples with standard supervision (S), terminal-only supervision (T), and nonterminal-only supervision (N), respectively.}

\label{tab:bert}

\end{center}
\end{table*}

The TOP dataset consists of trees where every token in the query is attached to either an intent (prefixed with IN:) or slot label (prefixed with SL:). Intents and slot labels can also attach to each other, forming compositional interpretations. 
We evaluate several models on the standard setup of the TOP dataset. We also devise new setups comparing the abilities of several models to learn from a smaller amount of fully annotated data, both with and without additional partially annotated data. Models are compared on exact match accuracy. Following \citet{rongali-etal-2020-generate}; \citet{einolghozati-etal-2019-improving}, and \citet{aghajanyan-etal-2020-conversational}, we filter out queries annotated as unsupported\footnote{We include results on the full set in Appendix~\ref{sec:full_data}.}, leaving 28414 \textit{train} examples and 8241 \textit{test} examples.

\subsection{Standard Supervision}
We use \emph{standard supervision} to refer to settings where all training examples contain a complete output tree. We also evaluate \emph{data efficiency}, by 
comparing the performance when training data is limited to 1\% or 10\% of the original dataset.

\subsection{Partial Supervision}
We use \emph{partial supervision} to refer to settings where we discard labels for certain nodes in the output trees of some or all training examples. Such partially annotated examples could arise in practice; for instance, when there is annotator disagreement on part of the output tree, or when changes to the set of possible slots or intents render parts of previously annotated trees obsolete. 

As semantic parsing datasets normally require expert annotators, extending fully annotated examples with additional partial annotation can be an effective strategy. For instance, \citet{choi-etal-2015-scalable} scaled their semantic parsing model with partial ontologies, and \citet{das-smith-2011-semi} used additional semi-supervised data for their frame semantic parsing model. We consider two types of partially annotated output trees described below.

\begin{figure}[t!]
\centering
\scalebox{0.85}{
\begin{tikzpicture}
\tikzset{level distance=25pt, level 1/.style={sibling distance=-30pt}}
\Tree[.\texttt{IN:GET\_DIRECTION}
    {\emph{directions} \emph{to}} ]
\begin{scope}[xshift=4cm]
\Tree[.\texttt{SL:ORGANIZER} {\emph{John}} ]
\end{scope}
\begin{scope}[yshift=-2cm]
\Tree[.\texttt{IN:FIND\_EVENT} {\emph{'s}} ]
\end{scope}
\begin{scope}[xshift=4cm,yshift=-2cm]
\Tree[.\texttt{SL:CATEGORY} {\emph{party}} ]
\end{scope}
\end{tikzpicture}
}
\caption{The TOP example from Figure~\ref{fig:top_example} with terminal-only supervision.}  
\label{fig:span_labels}
\end{figure}
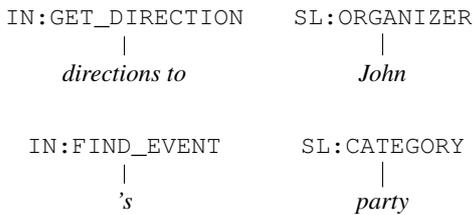

\paragraph{Terminal-only Supervision}
For this type of partial supervision, only the labels of each token (i.e., terminal) are preserved. See Figure~\ref{fig:span_labels} for an example. The label for each individual token is known, but the full set of intents and slots, and their tree structure, is unknown. This is similar to utilizing span labels that do not have full trees available.

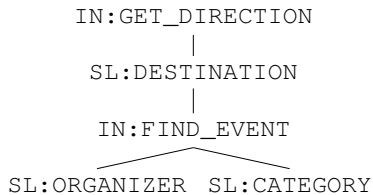
\begin{figure}[t!]
\centering
\scalebox{0.85}{
\begin{tikzpicture}
\tikzset{level distance=25pt, level 1/.style={sibling distance=-30pt}}
\Tree[.\texttt{IN:GET\_DIRECTION}
    [.\texttt{SL:DESTINATION}
        [.\texttt{IN:FIND\_EVENT}
            [.\texttt{SL:ORGANIZER} ]
            [.\texttt{SL:CATEGORY} ]
        ]
    ]
]
\end{tikzpicture}	
}
\caption{The TOP example from Figure~\ref{fig:top_example} with nonterminal-only supervision.}  
\label{fig:ungrounded_labels}
\end{figure}

\paragraph{Nonterminal-only Supervision}
For this type of partial supervision, token (i.e., terminal) labels are discarded. This is equivalent to deleting all of the token nodes from the tree. See Figure~\ref{fig:ungrounded_labels} for an example. This provides the opposite type of supervision as the terminal-only supervision case. The complete set of intents and slots and their tree structure is known, but their anchoring to the query text is unknown. For instance, if a query is known to have the same parse as a fully annotated query, its grounding may still be unknown.

\subsection{Results}

\paragraph{Comparisons with Fixed Encoder}
We first compare GBP with other methods using the same pre-trained encoder (BERT-Base; \citealt{devlin-etal-2019-bert}). We compare with a standard sequence decoder (a pointer-generator network; \citealt{vinyals2015pointer,see2017get}) implemented using a Transformer-based~\cite{vaswani-etal-2017-attention} decoder (\textsc{PtrGen}). We report the previous results from \citet{rongali-etal-2020-generate} and new results from an implementation based on that of ~\citet{suhr-etal-2020-exploring}, which provides a slightly stronger baseline. We also compare with the factored span parsing (FSP) approach of \citet{pasupat-etal-2019-span}. Notably, we report new results for FSP using a BERT-Base encoder, which are significantly stronger than previously published results which used GloVe~\cite{pennington-etal-2014-glove} embeddings (85.1\% vs.81.8\%).

Results can be found in Table~\ref{tab:bert}. We evaluate these models across both the standard and partial supervision settings. Notably, GBP can incorporate partial supervision in a straightforward way because scores for parse trees are factored over conditionally-independent scores for each edge. Training proceeds as described in Section~\ref{sec:model}; however, the loss from the edges that are not given by the example is masked. Additional training details can be found in Appendix~\ref{sec:training}. For PtrGen, each type of partial supervision is given a task-specific prefix; details are in Appendix~\ref{sec:t5}. Similar to GBP, FSP factors parse scores across local components, but also considers chains of length $>1$. Therefore, terminal-only supervision uses only length 1 chains; there is no trivial way to use nonterminal-only supervision without very substantial changes.

GBP is the highest-performing of the BERT-base models on the standard setup. Both GBP and FSP show better data efficiency than PtrGen. Only GBP appears to effectively benefit from partially annotated data in our experiments; the other models perform worse when incorporating this data.

\paragraph{Comparisons with Pretrained Decoders}
Recently, sequence-to-sequence models with pretrained decoders, such as BART~\cite{lewis-etal-2019-bart} and T5~\cite{raffel-etal-2020-exploring}, have demonstrated strong performance on a variety of tasks. Careful comparisons isolating the effects of model size and pretraining tasks are limited by the availability of pretrained checkpoints for such models. Regardless, we compare GBP (with BERT-Base) directly with such models. On the standard setting for TOP, ~\citet{aghajanyan-etal-2020-conversational} report SOTA performance with BART (87.1\%), outperforming GBP. We also report new results comparing GBP with T5 on both the standard supervision and partial supervision settings in Table~\ref{tab:GBPvT5}.
\begin{table}[t!]
\begin{center}
\scalebox{0.9}{
\begin{tabular}{lcccc}
\toprule
& \multicolumn{2}{c}{T5-Base} & \multicolumn{2}{c}{GBP (BERT-Base)} \\

 \cmidrule(lr){2-3} \cmidrule(lr){4-5}
 
S/T/N & Acc & Relative & Acc & Relative \\
\midrule
100 & \bf 86.26 & --- & 86.14 & --- \\
\midrule
10  & 77.70 & --- & \bf 79.43 & --- \\
10/90/0 & 83.44 & \phantom{0}7.39\% & \bf 84.55 & \phantom{0}6.45\% \\
10/0/90 & 81.63 & \phantom{0}5.06\% & \bf 84.14 & \phantom{0}5.93\%  \\
\midrule 
1 & 48.96 & --- & \bf 56.89 & --- \\
1/99/0 & 74.31 & 51.78\% & \bf 81.52 & 43.29\% \\
1/0/99 & 56.22 & 14.83\% & \bf 73.94 & 29.97\% \\
\midrule
2/49/49 & 82.00 & --- & \bf 85.01 & --- \\
0/50/50 & --- & --- & \bf 85.03 & --- \\
\bottomrule
\end{tabular}
}
\caption{Comparison of T5 and GBP on data efficiency and partial supervision. \emph{Relative} refers to the absolute increase in accuracy when incorporating partially-annotated examples compared to using only fully annotated data. S/T/N denotes the percentage of training examples with standard supervision (S), terminal-only supervision (T), and nonterminal-only supervision (N), respectively.
}
\label{tab:GBPvT5}

\end{center}
\end{table}

Notably, T5 is able to leverage partially-annotated examples much more effectively than \textsc{PtrGen}, which is also a Transformer-based sequence-to-sequence model but does not have a pretrained decoder. While T5 outperforms GBP on the standard setting, GBP outperforms T5 on the data efficiency and partial supervision settings.

\section{Related Work}
The most recent state of the art on TOP has focused on applying new methods of pretraining; (\citealt{rongali-etal-2020-generate}; \citealt{shao-etal-2020-graph}; \citealt{aghajanyan-etal-2020-conversational}) all use seq2seq methods, enhanced by better pretraining from BERT \citep{devlin-etal-2019-bert}, RoBERTa \citep{liu-etal-2019-roberta}, and BART \citep{lewis-etal-2020-bart} while using similar model architectures. In this work, we instead investigate the choice of decoder. While the FSP model \citep{pasupat-etal-2019-span} similarly uses a factored approach, its approach is more specific to TOP, as its trees must be projective and anchored to the input text.

In dependency parsing, the performance of graph-based and transition-based parsing is compared in both \citet{zhang-clark-2008-tale} and \citet{kulmizev-etal-2019-deep}. Graph-based parsing has also been used in AMR parsing \citep{zhang-etal-2019-amr}, which translates sentences into structured graph representations. Similar methods have also been used in semantic role labeling \citep{he-etal-2018-jointly}, which requires labeling arcs between text spans. This work is the first to adapt graph-based parsing to tree-like task-oriented semantic parses. 

\section{Conclusions}
We propose a novel framing of semantic parsing for TOP as a graph-based parsing task. We find that our proposed method is a competitive alternative to the standard paradigm of seq2seq models, especially when fully annotated data is limited and/or partially-annotated data is available.

\section*{Acknowledgements}
We thank Timothy Dozat, Terry Koo, and Philip Massey for their feedback as we developed this project and for their comments on earlier versions of the manuscript. We also thank the broader Google Language Research organization for their support of the project. Finally, we thank the anonymous reviewers for their helpful suggestions.  

\section*{Ethical Considerations}
We fine-tune all models using 32 Cloud TPU v3 cores\footnote{https://cloud.google.com/tpu/}. Additional training details are in Appendix~\ref{sec:t5} and Appendix~\ref{sec:training}. We reused existing pretrained checkpoints for both BERT and T5, reducing the resources needed to run experiments. Our evaluation focuses on the existing TOP dataset: the details of the collection can be found in \citet{gupta-etal-2018-semantic-parsing}. TOP is an English-only dataset, which limits our ability to claim that our findings generalize across languages.
A deployed dialog system has additional ethical considerations related to access, given their potential to make certain computational functions faster, easier, or more hands-free.

\bibliography{anthology,custom}

\begin{thebibliography}{33}
\expandafter\ifx\csname natexlab\endcsname\relax\def\natexlab#1{#1}\fi

\bibitem[{Aghajanyan et~al.(2020)Aghajanyan, Maillard, Shrivastava, Diedrick,
  Haeger, Li, Mehdad, Stoyanov, Kumar, Lewis, and
  Gupta}]{aghajanyan-etal-2020-conversational}
Armen Aghajanyan, Jean Maillard, Akshat Shrivastava, Keith Diedrick, Michael
  Haeger, Haoran Li, Yashar Mehdad, Veselin Stoyanov, Anuj Kumar, Mike Lewis,
  and Sonal Gupta. 2020.
\newblock \href {https://www.aclweb.org/anthology/2020.emnlp-main.408}
  {Conversational semantic parsing}.
\newblock In \emph{Proceedings of the 2020 Conference on Empirical Methods in
  Natural Language Processing (EMNLP)}, pages 5026--5035, Online. Association
  for Computational Linguistics.

\bibitem[{Choi et~al.(2015)Choi, Kwiatkowski, and
  Zettlemoyer}]{choi-etal-2015-scalable}
Eunsol Choi, Tom Kwiatkowski, and Luke Zettlemoyer. 2015.
\newblock \href {https://doi.org/10.3115/v1/P15-1127} {Scalable semantic
  parsing with partial ontologies}.
\newblock In \emph{Proceedings of the 53rd Annual Meeting of the Association
  for Computational Linguistics and the 7th International Joint Conference on
  Natural Language Processing (Volume 1: Long Papers)}, pages 1311--1320,
  Beijing, China. Association for Computational Linguistics.

\bibitem[{Chu and Liu(1965)}]{chu-1965-shortest}
Yoeng-Jin Chu and Tseng-Hong Liu. 1965.
\newblock On the shortest arborescence of a directed graph.
\newblock \emph{Scientia Sinica}, 14:1396--1400.

\bibitem[{Das and Smith(2011)}]{das-smith-2011-semi}
Dipanjan Das and Noah~A. Smith. 2011.
\newblock \href {https://www.aclweb.org/anthology/P11-1144} {Semi-supervised
  frame-semantic parsing for unknown predicates}.
\newblock In \emph{Proceedings of the 49th Annual Meeting of the Association
  for Computational Linguistics: Human Language Technologies}, pages
  1435--1444, Portland, Oregon, USA. Association for Computational Linguistics.

\bibitem[{Devlin et~al.(2019)Devlin, Chang, Lee, and
  Toutanova}]{devlin-etal-2019-bert}
Jacob Devlin, Ming-Wei Chang, Kenton Lee, and Kristina Toutanova. 2019.
\newblock \href {https://doi.org/10.18653/v1/N19-1423} {{BERT}: Pre-training of
  deep bidirectional transformers for language understanding}.
\newblock In \emph{Proceedings of the 2019 Conference of the North {A}merican
  Chapter of the Association for Computational Linguistics: Human Language
  Technologies, Volume 1 (Long and Short Papers)}, pages 4171--4186,
  Minneapolis, Minnesota. Association for Computational Linguistics.

\bibitem[{Dong and Lapata(2016)}]{dong-lapata-2016-language}
Li~Dong and Mirella Lapata. 2016.
\newblock \href {https://doi.org/10.18653/v1/P16-1004} {Language to logical
  form with neural attention}.
\newblock In \emph{Proceedings of the 54th Annual Meeting of the Association
  for Computational Linguistics (Volume 1: Long Papers)}, pages 33--43, Berlin,
  Germany. Association for Computational Linguistics.

\bibitem[{Dozat and Manning(2017)}]{dozat-manning-2017-biaffine}
Timothy Dozat and Christopher~D. Manning. 2017.
\newblock \href {https://openreview.net/forum?id=Hk95PK9le} {Deep biaffine
  attention for neural dependency parsing}.
\newblock In \emph{5th International Conference on Learning Representations,
  {ICLR} 2017, Toulon, France, April 24-26, 2017, Conference Track
  Proceedings}. OpenReview.net.

\bibitem[{Dyer et~al.(2016)Dyer, Kuncoro, Ballesteros, and
  Smith}]{dyer-etal-2016-recurrent}
Chris Dyer, Adhiguna Kuncoro, Miguel Ballesteros, and Noah~A. Smith. 2016.
\newblock \href {https://doi.org/10.18653/v1/N16-1024} {Recurrent neural
  network grammars}.
\newblock In \emph{Proceedings of the 2016 Conference of the North {A}merican
  Chapter of the Association for Computational Linguistics: Human Language
  Technologies}, pages 199--209, San Diego, California. Association for
  Computational Linguistics.

\bibitem[{Edmonds(1965)}]{edmonds1965paths}
Jack Edmonds. 1965.
\newblock Paths, trees, and flowers.
\newblock \emph{Canadian Journal of mathematics}, 17:449--467.

\bibitem[{Einolghozati et~al.(2018)Einolghozati, Pasupat, Gupta, Shah, Mohit,
  Lewis, and Zettlemoyer}]{einolghozati-etal-2019-improving}
Arash Einolghozati, Panupong Pasupat, Sonal Gupta, Rushin Shah, Mrinal Mohit,
  Mike Lewis, and Luke Zettlemoyer. 2018.
\newblock Improving semantic parsing for task oriented dialog.
\newblock In \emph{Conversational AI Workshop at NeurIPS}.

\bibitem[{Gupta et~al.(2018)Gupta, Shah, Mohit, Kumar, and
  Lewis}]{gupta-etal-2018-semantic-parsing}
Sonal Gupta, Rushin Shah, Mrinal Mohit, Anuj Kumar, and Mike Lewis. 2018.
\newblock \href {https://doi.org/10.18653/v1/D18-1300} {Semantic parsing for
  task oriented dialog using hierarchical representations}.
\newblock In \emph{Proceedings of the 2018 Conference on Empirical Methods in
  Natural Language Processing}, pages 2787--2792, Brussels, Belgium.
  Association for Computational Linguistics.

\bibitem[{He et~al.(2018)He, Lee, Levy, and Zettlemoyer}]{he-etal-2018-jointly}
Luheng He, Kenton Lee, Omer Levy, and Luke Zettlemoyer. 2018.
\newblock \href {https://doi.org/10.18653/v1/P18-2058} {Jointly predicting
  predicates and arguments in neural semantic role labeling}.
\newblock In \emph{Proceedings of the 56th Annual Meeting of the Association
  for Computational Linguistics (Volume 2: Short Papers)}, pages 364--369,
  Melbourne, Australia. Association for Computational Linguistics.

\bibitem[{Jia and Liang(2016)}]{jia-liang-2016-data}
Robin Jia and Percy Liang. 2016.
\newblock \href {https://doi.org/10.18653/v1/P16-1002} {Data recombination for
  neural semantic parsing}.
\newblock In \emph{Proceedings of the 54th Annual Meeting of the Association
  for Computational Linguistics (Volume 1: Long Papers)}, pages 12--22, Berlin,
  Germany. Association for Computational Linguistics.

\bibitem[{Kiperwasser and Goldberg(2016)}]{kiperwasser2016simple}
Eliyahu Kiperwasser and Yoav Goldberg. 2016.
\newblock Simple and accurate dependency parsing using bidirectional lstm
  feature representations.
\newblock \emph{Transactions of the Association for Computational Linguistics},
  4:313--327.

\bibitem[{Kulmizev et~al.(2019)Kulmizev, de~Lhoneux, Gontrum, Fano, and
  Nivre}]{kulmizev-etal-2019-deep}
Artur Kulmizev, Miryam de~Lhoneux, Johannes Gontrum, Elena Fano, and Joakim
  Nivre. 2019.
\newblock \href {https://doi.org/10.18653/v1/D19-1277} {Deep contextualized
  word embeddings in transition-based and graph-based dependency parsing - a
  tale of two parsers revisited}.
\newblock In \emph{Proceedings of the 2019 Conference on Empirical Methods in
  Natural Language Processing and the 9th International Joint Conference on
  Natural Language Processing (EMNLP-IJCNLP)}, pages 2755--2768, Hong Kong,
  China. Association for Computational Linguistics.

\bibitem[{Lewis et~al.(2019)Lewis, Liu, Goyal, Ghazvininejad, Mohamed, Levy,
  Stoyanov, and Zettlemoyer}]{lewis-etal-2019-bart}
Mike Lewis, Yinhan Liu, Naman Goyal, Marjan Ghazvininejad, Abdelrahman Mohamed,
  Omer Levy, Ves Stoyanov, and Luke Zettlemoyer. 2019.
\newblock Bart: Denoising sequence-to-sequence pre-training for natural
  language generation, translation, and comprehension.
\newblock \emph{arXiv preprint arXiv:1910.13461}.

\bibitem[{Lewis et~al.(2020)Lewis, Liu, Goyal, Ghazvininejad, Mohamed, Levy,
  Stoyanov, and Zettlemoyer}]{lewis-etal-2020-bart}
Mike Lewis, Yinhan Liu, Naman Goyal, Marjan Ghazvininejad, Abdelrahman Mohamed,
  Omer Levy, Veselin Stoyanov, and Luke Zettlemoyer. 2020.
\newblock \href {https://doi.org/10.18653/v1/2020.acl-main.703} {{BART}:
  Denoising sequence-to-sequence pre-training for natural language generation,
  translation, and comprehension}.
\newblock In \emph{Proceedings of the 58th Annual Meeting of the Association
  for Computational Linguistics}, pages 7871--7880, Online. Association for
  Computational Linguistics.

\bibitem[{Liu et~al.(2019)Liu, Ott, Goyal, Du, Joshi, Chen, Levy, Lewis,
  Zettlemoyer, and Stoyanov}]{liu-etal-2019-roberta}
Yinhan Liu, Myle Ott, Naman Goyal, Jingfei Du, Mandar Joshi, Danqi Chen, Omer
  Levy, Mike Lewis, Luke Zettlemoyer, and Veselin Stoyanov. 2019.
\newblock Roberta: A robustly optimized bert pretraining approach.
\newblock \emph{arXiv preprint arXiv:1907.11692}.

\bibitem[{McDonald et~al.(2005)McDonald, Crammer, and
  Pereira}]{mcdonald2005online}
Ryan McDonald, Koby Crammer, and Fernando Pereira. 2005.
\newblock Online large-margin training of dependency parsers.
\newblock In \emph{Proceedings of the 43rd Annual Meeting of the Association
  for Computational Linguistics (ACL’05)}, pages 91--98.

\bibitem[{Pasupat et~al.(2019)Pasupat, Gupta, Mandyam, Shah, Lewis, and
  Zettlemoyer}]{pasupat-etal-2019-span}
Panupong Pasupat, Sonal Gupta, Karishma Mandyam, Rushin Shah, Mike Lewis, and
  Luke Zettlemoyer. 2019.
\newblock \href {https://doi.org/10.18653/v1/D19-1163} {Span-based hierarchical
  semantic parsing for task-oriented dialog}.
\newblock In \emph{Proceedings of the 2019 Conference on Empirical Methods in
  Natural Language Processing and the 9th International Joint Conference on
  Natural Language Processing (EMNLP-IJCNLP)}, pages 1520--1526, Hong Kong,
  China. Association for Computational Linguistics.

\bibitem[{Pennington et~al.(2014)Pennington, Socher, and
  Manning}]{pennington-etal-2014-glove}
Jeffrey Pennington, Richard Socher, and Christopher Manning. 2014.
\newblock \href {https://doi.org/10.3115/v1/D14-1162} {{G}lo{V}e: Global
  vectors for word representation}.
\newblock In \emph{Proceedings of the 2014 Conference on Empirical Methods in
  Natural Language Processing ({EMNLP})}, pages 1532--1543, Doha, Qatar.
  Association for Computational Linguistics.

\bibitem[{Raffel et~al.(2020)Raffel, Shazeer, Roberts, Lee, Narang, Matena,
  Zhou, Li, and Liu}]{raffel-etal-2020-exploring}
Colin Raffel, Noam Shazeer, Adam Roberts, Katherine Lee, Sharan Narang, Michael
  Matena, Yanqi Zhou, Wei Li, and Peter~J Liu. 2020.
\newblock Exploring the limits of transfer learning with a unified text-to-text
  transformer.
\newblock \emph{Journal of Machine Learning Research}, 21(140):1--67.

\bibitem[{Rongali et~al.(2020)Rongali, Soldaini, Monti, and
  Hamza}]{rongali-etal-2020-generate}
Subendhu Rongali, Luca Soldaini, Emilio Monti, and Wael Hamza. 2020.
\newblock Don’t parse, generate! a sequence to sequence architecture for
  task-oriented semantic parsing.
\newblock In \emph{Proceedings of The Web Conference 2020}, pages 2962--2968.

\bibitem[{See et~al.(2017)See, Liu, and Manning}]{see2017get}
Abigail See, Peter~J Liu, and Christopher~D Manning. 2017.
\newblock Get to the point: Summarization with pointer-generator networks.
\newblock \emph{arXiv preprint arXiv:1704.04368}.

\bibitem[{Shao et~al.(2020)Shao, Gong, Qi, Cao, Ji, and
  Lin}]{shao-etal-2020-graph}
Bo~Shao, Yeyun Gong, Weizhen Qi, Guihong Cao, Jianshu Ji, and Xiaola Lin. 2020.
\newblock \href {https://doi.org/10.1609/aaai.v34i05.6408} {Graph-based
  transformer with cross-candidate verification for semantic parsing}.
\newblock \emph{Proceedings of the AAAI Conference on Artificial Intelligence},
  34(05):8807--8814.

\bibitem[{Shaw et~al.(2019)Shaw, Massey, Chen, Piccinno, and
  Altun}]{shaw-etal-2019-generating}
Peter Shaw, Philip Massey, Angelica Chen, Francesco Piccinno, and Yasemin
  Altun. 2019.
\newblock \href {https://doi.org/10.18653/v1/P19-1010} {Generating logical
  forms from graph representations of text and entities}.
\newblock In \emph{Proceedings of the 57th Annual Meeting of the Association
  for Computational Linguistics}, pages 95--106, Florence, Italy. Association
  for Computational Linguistics.

\bibitem[{Suhr et~al.(2020)Suhr, Chang, Shaw, and
  Lee}]{suhr-etal-2020-exploring}
Alane Suhr, Ming-Wei Chang, Peter Shaw, and Kenton Lee. 2020.
\newblock \href {https://doi.org/10.18653/v1/2020.acl-main.742} {Exploring
  unexplored generalization challenges for cross-database semantic parsing}.
\newblock In \emph{Proceedings of the 58th Annual Meeting of the Association
  for Computational Linguistics}, pages 8372--8388, Online. Association for
  Computational Linguistics.

\bibitem[{Vaswani et~al.(2017)Vaswani, Shazeer, Parmar, Uszkoreit, Jones,
  Gomez, Kaiser, and Polosukhin}]{vaswani-etal-2017-attention}
Ashish Vaswani, Noam Shazeer, Niki Parmar, Jakob Uszkoreit, Llion Jones,
  Aidan~N Gomez, {\L}ukasz Kaiser, and Illia Polosukhin. 2017.
\newblock Attention is all you need.
\newblock In \emph{Advances in neural information processing systems}, pages
  5998--6008.

\bibitem[{Vinyals et~al.(2015)Vinyals, Fortunato, and
  Jaitly}]{vinyals2015pointer}
Oriol Vinyals, Meire Fortunato, and Navdeep Jaitly. 2015.
\newblock Pointer networks.
\newblock In \emph{Proceedings of the 28th International Conference on Neural
  Information Processing Systems-Volume 2}, pages 2692--2700.

\bibitem[{Wang et~al.(2019{\natexlab{a}})Wang, Shin, Liu, Polozov, and
  Richardson}]{wang-etal-2019-rat}
Bailin Wang, Richard Shin, Xiaodong Liu, Oleksandr Polozov, and Matthew
  Richardson. 2019{\natexlab{a}}.
\newblock Rat-sql: Relation-aware schema encoding and linking for text-to-sql
  parsers.
\newblock \emph{arXiv preprint arXiv:1911.04942}.

\bibitem[{Wang et~al.(2019{\natexlab{b}})Wang, Tian, He, Qin, Zhai, and
  Liu}]{wang-etal-2019-non}
Yiren Wang, Fei Tian, Di~He, Tao Qin, ChengXiang Zhai, and Tie-Yan Liu.
  2019{\natexlab{b}}.
\newblock Non-autoregressive machine translation with auxiliary regularization.
\newblock In \emph{Proceedings of the AAAI Conference on Artificial
  Intelligence}, volume~33, pages 5377--5384.

\bibitem[{Zhang et~al.(2019)Zhang, Ma, Duh, and
  Van~Durme}]{zhang-etal-2019-amr}
Sheng Zhang, Xutai Ma, Kevin Duh, and Benjamin Van~Durme. 2019.
\newblock \href {https://doi.org/10.18653/v1/P19-1009} {{AMR} parsing as
  sequence-to-graph transduction}.
\newblock In \emph{Proceedings of the 57th Annual Meeting of the Association
  for Computational Linguistics}, pages 80--94, Florence, Italy. Association
  for Computational Linguistics.

\bibitem[{Zhang and Clark(2008)}]{zhang-clark-2008-tale}
Yue Zhang and Stephen Clark. 2008.
\newblock \href {https://www.aclweb.org/anthology/D08-1059} {A tale of two
  parsers: {I}nvestigating and combining graph-based and transition-based
  dependency parsing}.
\newblock In \emph{Proceedings of the 2008 Conference on Empirical Methods in
  Natural Language Processing}, pages 562--571, Honolulu, Hawaii. Association
  for Computational Linguistics.

\end{thebibliography}
\bibliographystyle{acl_natbib}

\newpage
\appendix

\section{Baseline Model Training Details}
\label{sec:t5}
\subsection{T5 Training Details}
We use the base version of the T5.1.1 model (220M parameters)\footnote{https://github.com/google-research/text-to-text-transfer-transformer} for finetuning with default learning rates. We also experimented with T5-large in preliminary experiments but did not observe visible difference in performances. The multitask experiments are finetuned for 5000 steps, and the rest are finetuned for 2000 steps, all with $\text{batch\_size}=4096$. We use greedy decoding at inference time. All experiments are ran once. 

We use 32 Cloud TPU v3 cores for training and 8 TPU cores for inference. Training each model takes about 4 hours, and inference takes about 2 minutes on the entire TOP test set. 

For the multitask experiments, we follow~\citet{raffel-etal-2020-exploring} by appending task prefixes to the input sequence. Specifically, we used ``span:'' for examples with terminal-only partial supervision, ``ungrounded:'' for examples with nonterminal-only partial supervision, and ``full:'' for examples with full supervision. 

\subsection{PtrGen Training Details}
Following \citet{suhr-etal-2020-exploring}, we started with the hyperparameters of \cite{shaw-etal-2019-generating}. We then tuned the learning rate over $3$ runs to be $1e^{-4}$. We use a BERT-Base encoder and the Transformer decoder consists of $4$ layers with $8$ attention heads, $64$ dimensions, and $256$ feed-forward hidden layer dimensions. The model is trained for $30K$ steps, with the pre-trained BERT parameters frozen for the first $4K$ steps. Task prefixes are prepended in the same manner as T5.

\subsection{FSP Training Details}
Hyperparameters were reused from \citet{pasupat-etal-2019-span}, except for the initial learning rate which changes to 0.00001 to make it more suitable for fine-tuning BERT. For partially supervised examples, we only define the loss on the spans that are labeled in the example, and re-weight the loss by a factor of 0.01 (tuned on development data).

\section{GBP Training Details}
\label{sec:training}
The model takes BERT wordpieces from a publicly available BERT-base checkpoint\footnote{\href{https://huggingface.co/google/bert_uncased_L-12_H-768_A-12}{https://huggingface.co/google/bert\_uncased\_L-12\_H-768\_A-12}} \citep{devlin-etal-2019-bert}. Intents have slots have randomly initialized 768-dimensional embeddings. The Transformer encoder uses 4 layers of cross-attention \citep{vaswani-etal-2017-attention} with 4 attention heads and 768 dims and a dropout rate of 0.3. 

We use a hidden size of 1024 for computing edge scores similarly to \citet{dozat-manning-2017-biaffine}.
Cross entropy loss is minimized with the optimizer described in \citet{devlin-etal-2019-bert}. 

For partial supervision experiments, the loss is masked for unsupervised edges.

The model is trained over 20000 steps with a learning rate of 0.0001 and 2000 warmup steps. All hyperparameters are chosen based on validation set exact match accuracy performed by a grid search. 
BERT-base has approximately 110M parameters, and GBSP introduces approximately 13M additional parameters, for a total of approximately 123M parameters.
Note that larger versions of BERT did not lead to performance improvements in our experiments.

Note a comparison on validation performance can be found in Table~\ref{tab:validation} (the validation set without unsupported has 4032 examples). All tested values for hyperpameters can be found in Table~\ref{tab:hparams}. 
We estimate approximately 1,000 total training runs during the development cycle.
After tuning hyperparameters on the full set, no re-tuning occurred: partial supervision and data efficiency experiments used the same setup. Model training takes approximately 45 minutes.

\begin{table}[t!]
\begin{center}
\scalebox{0.9}{
\begin{tabular}{lc}
\toprule
Model & Acc \\
\midrule
\textsc{T5} \citep{raffel-etal-2020-exploring} & 85.22 \\
\midrule
\textsc{GBP} (Proposed) & 85.17  \\
\textsc{FSP} \citep{pasupat-etal-2019-span} & 84.53 \\
\textsc{PtrGen} (Ours) & 85.08 \\
\bottomrule
\end{tabular}
}
\caption{Accuracy results for the TOP dataset evaluated on the validation set. Note all models besides T5 are initialized from BERT-Base}
\label{tab:validation}

\end{center}
\end{table}
\begin{table}[t!]
\begin{center}
\scalebox{0.8}{
\begin{tabular}{l|cccc}
\toprule
Parameter & Start & End & Incr. & Num \\ 
\midrule
Node Encoder Dim & 128 & 2048 & x2 & 5 \\
Biaffine Hidden Dim & 128 & 2048 & x2 & 5 \\
Learning Rate & 0.0001 & 0.01 & x10 & 3 \\
Number of Heads & 2 & 8 & x2 & 3 \\
Warmup & 2000 & 4000 & x2 & 2 \\
Number of Layers & 1 & 8 & x2 & 4 \\
\bottomrule
\end{tabular}
}
\caption{Hyperparameter sweep for GBSP.}
\label{tab:hparams}

\end{center}
\end{table}

\section{Repeated Nodes}
\label{sec:repeated}
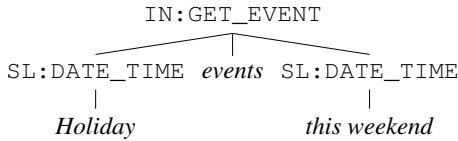
\begin{figure}[t!]
\centering
\scalebox{0.85}{
\begin{tikzpicture}
\tikzset{level distance=25pt, level 1/.style={sibling distance=0pt}}
\Tree[.\texttt{IN:GET\_EVENT}
 [.\texttt{SL:DATE\_TIME} \emph{Holiday} ]
 \emph{events}
  [.\texttt{SL:DATE\_TIME} \emph{this weekend} ]
]
\end{tikzpicture}
}
\caption{Example TOP tree with two occurrences of  \texttt{SL:DATE\_TIME}. When mapping from TOP trees to the parse trees predicted by our model, each instance of \texttt{SL:DATE\_TIME} is assigned an index based it its preorder position in the TOP tree.}
\label{fig:repeated_example}
\end{figure}

See Figure~\ref{fig:repeated_example} for an example of a TOP tree with repeated nodes.

We chose to pad occurrences based on the observation that certain nodes can occur more times than they do in the training set. About half of the nodes only ever occur once. On the validation set, 2 additional replications was the highest value before performance degraded.

There are many alternatives to our handling of repeated nodes. For instance, \citet{zhang-etal-2019-amr} had a slightly different task, but we could have adopted their approach of generating the node set auto-regressively. Unfortunately, this would have complicated our method of partial supervision. Another method would be to use a fixed number of duplications: this worked slightly worse in practice, based on validation set performance. Alternatively, the model could have learned a regression, which has been used in non-autoregressive machine translation (e.g., \citealt{wang-etal-2019-non}). We leave trying such an approach to future work.

\section{Full Data}
\label{sec:full_data}
Results on the full dataset (including unsupported intents) can be found in Table~\ref{tab:full}. Note that most recent work does not report on this setting. For this setting, we train on every example in \textit{train} (31279 examples) and evaluate on every example in \textit{test} (9042 examples). Note that the full dataset is available at \href{http://fb.me/semanticparsingdialo}{http://fb.me/semanticparsingdialog}; unsupported intents were excluded manually with string matching.

\begin{table}[t!]
\begin{center}
\scalebox{0.9}{
\begin{tabular}{lc}
\toprule
Model & Acc \\
\midrule
\textsc{RNNG} \citep{dyer-etal-2016-recurrent} & 78.51 \\
\textsc{GTCV} \citep{shao-etal-2020-graph} & 82.51 \\
\midrule
\textsc{T5} \citep{raffel-etal-2020-exploring} & 84.11 \\
\textsc{GBP} (Proposed) & 83.31  \\
\bottomrule
\end{tabular}
}
\caption{Accuracy results for the TOP dataset evaluated on all test examples, including those with unsupported intents.}
\label{tab:full}

\end{center}
\end{table}
\end{document}